\documentclass[letterpaper, 10 pt, conference]{ieeeconf}

\IEEEoverridecommandlockouts
\usepackage[utf8]{inputenc}
\usepackage{amsmath}
\usepackage{amssymb}
\usepackage{graphicx}
\usepackage{placeins}
\usepackage{booktabs}
\usepackage{capt-of}
\usepackage{cite}
\usepackage{microtype}
\makeatletter
\let\NAT@parse\undefined
\makeatother
\usepackage{hyperref}
\DeclareUnicodeCharacter{201C}{``}
\DeclareUnicodeCharacter{201D}{''}
\DeclareUnicodeCharacter{2018}{`}
\DeclareUnicodeCharacter{2019}{'}
\DeclareUnicodeCharacter{2013}{--}
\DeclareUnicodeCharacter{2014}{---}
\DeclareUnicodeCharacter{2191}{$\uparrow$}
\DeclareUnicodeCharacter{2193}{$\downarrow$}
\DeclareUnicodeCharacter{2713}{$\checkmark$}
\DeclareUnicodeCharacter{2717}{$\times$}

\title{\LARGE \bf
CoFiNav: VLM-Guided Coarse-to-Fine Semantic Navigation for Open-Ended Last-Mile UAV Delivery
}

\author{Xuanpu Wang$^{1}$, Shun Hu$^{1}$, Bing Li$^{1}$, and Rongqing Zhang$^{2}$%
\thanks{$^{1}$ Xuanpu Wang, Shun Hu, and Bing Li are with Tongji University. \{2451267,2433295,lizi\}@tongji.edu.cn}%
\thanks{$^{2}$ Rongqing Zhang is with The Hong Kong University of Science and Technology (Guangzhou). rongqingz@tongji.edu.cn}%
}

\begin{document}

\maketitle
\thispagestyle{empty}
\pagestyle{empty}
\setlength{\parskip}{0pt}

\begin{abstract}
Unmanned aerial vehicle (UAV) delivery is an important application of the low-altitude economy, enhancing the flexibility and accessibility of last-mile logistics. Existing UAV delivery methods generally rely on precise terminal coordinates, predefined terminal specifications, or task-specific localization procedures, limiting the scope and flexibility of delivery. As UAV logistics becomes more intelligent, open-ended semantic descriptions offer a new paradigm for specifying terminal destinations. However, conventional aerial navigation methods have limited ability to generalize to delivery scenarios that require open-vocabulary terminal navigation. To enable this broader delivery paradigm, we propose CoFiNav, a VLM-guided coarse-to-fine UAV semantic navigation framework that progressively grounds relation-defined terminal destinations through hierarchical semantic search and active 3D observation. Given a task description, a coarse prior map from OpenStreetMap (OSM), and accumulated observations, CoFiNav uses a vision-language model (VLM) to determine which unresolved entity or spatial relation should be verified next. A geometry-aware module converts the semantic observation request into feasible viewpoints using RGB images and LiDAR, while conventional motion planning handles safe execution. Experiments across multiple urban environments show that CoFiNav outperforms representative adapted baselines in success rate while requiring shorter flight paths and fewer observations.
\end{abstract}

\section{INTRODUCTION}

\begin{figure}[t]
    \centering
    \includegraphics[width=0.88\columnwidth]{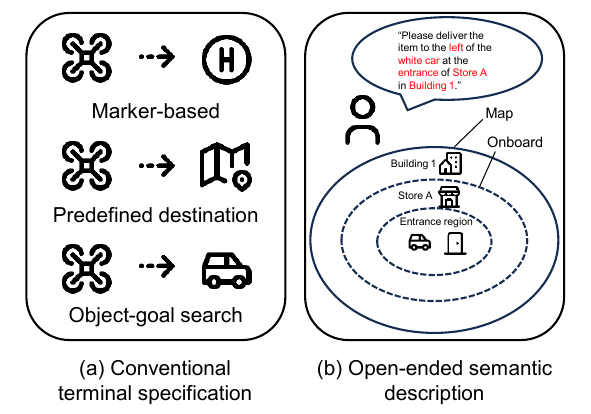}
    \caption{Comparison of terminal destination specifications. (a) Conventional settings specify an explicit terminal. (b) Open-ended semantic descriptions use hierarchical references and spatial relations: the map provides the task-building footprint, while finer cues are resolved onboard.}
    \label{fig:task_setting}
\end{figure}

As a representative application of the low-altitude economy, unmanned aerial vehicle (UAV) delivery overcomes the constraints of ground road networks and substantially improves the flexibility and accessibility of last-mile logistics. As UAV logistics advances toward more flexible and intelligent delivery scenarios, moving beyond fixed addresses to open-ended semantic descriptions offers a new paradigm for specifying terminal destinations, as illustrated in Fig.~\ref{fig:task_setting}. However, existing UAV logistics research focuses primarily on planning and scheduling, generally assuming that precise delivery destinations are known. In practice, the local environmental details and terminal information needed to determine the final destination may be unavailable in advance. Therefore, task-relevant perception and reasoning must rely on onboard observations. Few studies consider delivery in map-free terminal scenarios, but typically rely on predefined specifications, external markers, and task-specific localization procedures to determine the final destination \cite{UrbanLastMile,DoorYard,RSSS,LogisticsVLN}. These studies restrict the scope of delivery and do not fully exploit the flexibility and intelligence of UAV delivery. How to autonomously navigate and determine the destination during approach from open-ended descriptions remains an open challenge.

Such a process requires aerial navigation capabilities in complex urban environments. Conventional aerial navigation methods without foundation models have limited generalization to real-world delivery scenarios that require open-vocabulary terminal navigation. Recent studies have therefore focused on zero-shot navigation, using large language models (LLMs) and vision-language models (VLMs) for language parsing, spatial reasoning, and action generation \cite{GeoNav}. However, they rely heavily on action requirements specified in language instructions, limiting autonomous decision-making \cite{CityNavAgent,LookasideVLN}. At the terminal stage, fine-grained cues are often not reliably observable from conventional aerial views due to scale and occlusion. Although semantic and spatial reasoning has advanced in aerial navigation, the ability to actively adjust altitude or viewing direction to resolve task-relevant cues remains insufficiently studied \cite{GeoNav}. Some studies focus on autonomous exploration and navigation in unknown environments, but they only locate specified objects and generally do not search over relations among multiple semantic entities \cite{CityAVOS,AirHunt}, while open-ended semantic destinations are often defined by hierarchical relations among multiple entities. Accordingly, these limitations highlight the absence of a closed-loop process in which unresolved semantic structure guides the acquisition of task-relevant observations and progressively grounds a relation-defined terminal destination.

We therefore propose CoFiNav, a VLM-guided coarse-to-fine UAV semantic navigation framework that maintains and advances a hierarchical semantic search process. Given the task description, a coarse prior map from OpenStreetMap (OSM), and accumulated observations, the VLM reasons about which unresolved entity or spatial relation should be verified next. The resulting semantic observation request specifies what needs to be observed, while a geometry-aware module grounds it into where it can be reliably observed using RGB images and LiDAR. Candidate viewpoints are evaluated according to visibility, projected scale, viewing angle, occlusion, and contextual coverage, after which conventional motion planning handles safe trajectory execution. This separation allows foundation-model reasoning to operate on task semantics while leaving metric geometry and flight safety to deterministic components, enabling a closed-loop progression from coarse geographic context to fine-grained target grounding through active 3D observation. Experiments show that CoFiNav achieves the highest success rate among adapted baselines while requiring shorter flight paths and fewer semantic observations.

Our main contributions are summarized as follows:

\begin{itemize}

\item Coarse-to-fine UAV semantic navigation for a new paradigm of last-mile delivery. Existing UAV delivery systems largely rely on predefined terminal specifications or task-specific localization procedures, limiting their applicability to open-ended semantic destinations. To address this limitation, we formulate terminal UAV navigation as a progressive grounding problem across multiple spatial scales, connecting large-scale aerial navigation with fine-grained terminal grounding during approach.

\item VLM-guided hierarchical semantic search. Relational terminal goals cannot generally be reduced to an isolated search target. To account for their semantic dependencies, we maintain a dynamic hierarchical search state over unresolved entities, attributes, and spatial relations, and use VLM reasoning to prioritize which element should be verified next, allowing the search to evolve with accumulated observations.

\item Geometry-grounded active 3D observation. Fixed aerial observations may provide insufficient information for fine-grained grounding because of viewing direction, scale, or occlusion. To overcome this limitation, we develop an observation-grounding mechanism that converts semantic observation requests into feasible 3D viewpoints, enabling the UAV to actively adjust altitude, distance, and viewing direction according to the current observation requirement.

\item Realistic urban simulation platform for last-mile UAV delivery. Existing UAV simulation environments provide
large-scale scenes with limited fine-grained detail and
restricted 3D viewpoint variation. Thus, we construct a simulation platform based on UrbanScene3D, aligned with real-world OSM priors and enriched with fine-grained semantic details and diverse delivery tasks. On this platform, we evaluate CoFiNav across four typical environments and compare it against representative adapted baselines. CoFiNav achieves an 81.7\% success rate with only 6.6 semantic observations on average.

\end{itemize}
\section{RELATED WORK}

\indent\textbf{Autonomous Urban UAV Delivery.}\ UAV delivery enables the autonomous aerial transportation of packages from an origin to a designated destination \cite{AerialRoboticsReview}. Existing research has primarily focused on long-range routing and scheduling over known maps \cite{DroneRoutingReview}, as well as multi-robot coordination \cite{TruckDroneRoutingReview}. These studies generally assume that the task is completed once the UAV reaches the designated location, leaving fine-grained map-free localization in the last mile largely unaddressed. Some studies have explored map-free last-mile delivery. Early systems navigate to an approximate location and use visual markers to determine the final drop-off point \cite{UrbanLastMile}, while marker-free approaches localize predefined semantic destinations such as doors or yards \cite{DoorYard,SSS,RSSS}. LogisticsVLN adopts a staged procedure involving floor localization, target-window search, and final approach for window-level delivery \cite{LogisticsVLN}. These formulations remain largely tied to predefined destination types, task-specific localization procedures, or predetermined exploration structures, limiting the scope and autonomy of delivery. Some ground-robot delivery studies have begun to consider addresses specified by semantic descriptions. OPEN addresses last-mile address navigation using coarse OSM information and local visual exploration when the available geographic prior is insufficient \cite{OpenBench}. However, it uniformly samples viewpoints around the target building and sequentially inspects them with a VLM, making the local search largely coverage-driven. More general terminal instructions may require the destination to be progressively established from unresolved semantic relations as the required observations are acquired during approach.

\indent\textbf{UAV Semantic Navigation.}\ UAV semantic navigation incorporates high-level semantic information about the environment and navigation goals into aerial perception and decision-making \cite{SemanticNavSurvey}. Conventional methods without foundation models are typically trained on paired trajectories and language instructions. AerialVLN \cite{AerialVLN}, CityNav \cite{CityNav}, and OpenFly \cite{OpenFly} extend aerial navigation toward geographic-semantic representations and larger-scale trajectory datasets. However, such methods remain difficult to generalize to real-world delivery scenarios that require open-vocabulary terminal navigation. To reduce this dependence on task-specific training, recent studies introduce large language and vision-language models to improve semantic and spatial reasoning \cite{OpenUAV,STMR,CityNavAgent,LookasideVLN}. CityNavAgent uses LLM-based hierarchical semantic planning to decompose long-horizon aerial navigation into semantic subgoals \cite{CityNavAgent}, and LookasideVLN models landmark-direction relationships for spatial reasoning \cite{LookasideVLN}. GeoNav combines multimodal LLM reasoning with global geographic priors and local hierarchical scene representations for zero-shot language-goal navigation \cite{GeoNav}. However, its experiments are conducted at fixed flight altitudes with predominantly top-down observations, and the UAV's three-dimensional mobility is not explicitly exploited for active viewpoint adjustment. These studies depend on long-horizon human language instructions and lack autonomous active exploration in unknown environments. Some methods incorporate semantic information into autonomous exploration and navigation \cite{CityAVOS,AirHunt} through semantic maps, frontier selection, and persistent scene representations \cite{SemExp,FSE,APEX,RAVEN}. VLFM constructs vision-language frontier maps and uses language-conditioned semantic relevance to prioritize exploration frontiers for zero-shot semantic navigation \cite{VLFM}. However, these approaches do not explicitly represent the relational dependencies required when the final goal must be established from multiple semantically related entities.

\section{PROBLEM FORMULATION}

At the beginning of a task, the UAV is provided with \(\mathcal{I}_0=(L,M_0^{\mathrm{prior}},\xi_0)\), where \(L\) denotes the natural-language task instruction, \(M_0^{\mathrm{prior}}\) denotes a coarse prior map of the environment, and \(\xi_0\) denotes the initial UAV pose in a unified world coordinate frame. The prior map mainly contains coarse spatial information such as building footprints and roads, while finer-grained information must be resolved during task execution.

During execution, the \(k\)-th observation is denoted as \(O_k=(I_k,P_k,\xi_k)\), where \(I_k\) is the RGB image, \(P_k\) is the 3D LiDAR observation, and \(\xi_k\) is the corresponding UAV pose. Based on the initial task input \(\mathcal{I}_0\) and the accumulated observation history \(H_k=(O_0,\ldots,O_k)\), the system is required to progressively ground the terminal delivery location specified by the instruction, represented as \(\hat{\mathbf{p}}\in\mathbb{R}^3\) in the unified world coordinate frame. For each task, a ground-truth valid delivery region \(\mathcal{G}^\ast\subset\mathbb{R}^3\) is manually annotated during task construction to represent the set of feasible terminal positions that satisfy the complete semantic and spatial constraints expressed in the instruction. A task is considered successful if \(\hat{\mathbf{p}}\in\mathcal{G}^\ast\).

\section{COFINAV FRAMEWORK}

\subsection{Method Overview}

\begin{figure*}[t]
    \centering
    \includegraphics[width=\textwidth]{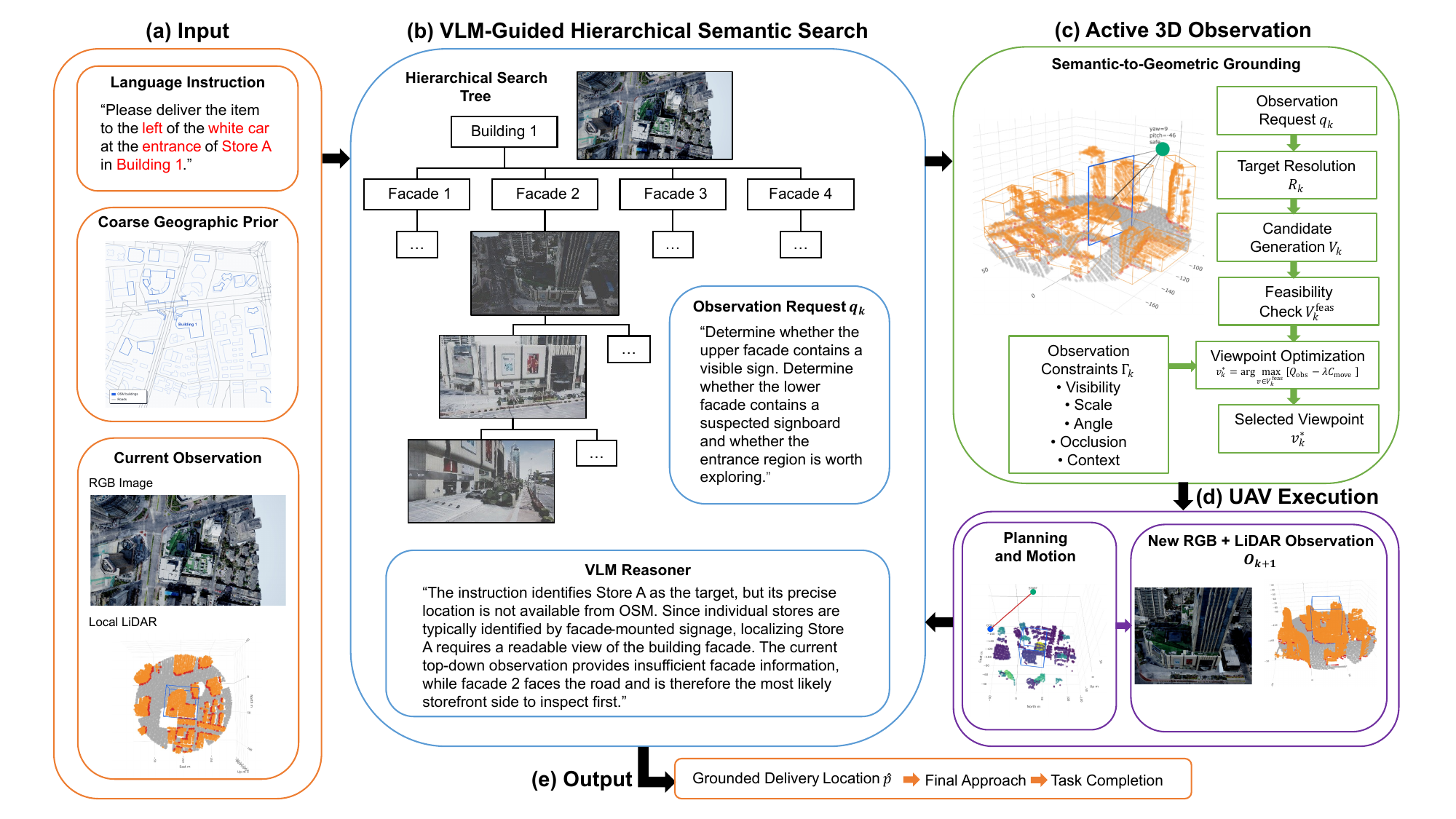}
    \caption{Overview of the proposed CoFiNav framework. (a) The language instruction, coarse geographic prior, and current RGB/LiDAR observation provide the task context. (b) The VLM advances hierarchical semantic search and generates the next observation request. (c) The request is grounded into a feasible 3D viewpoint through semantic-to-geometric grounding and task-conditioned viewpoint optimization. (d) The UAV executes the planned motion and feeds the newly acquired RGB/LiDAR observation back to the search loop. (e) Once the terminal location is grounded, the estimated delivery location $\hat{\mathbf{p}}$ is passed to final approach and task completion.}
    \label{fig:overview}
\end{figure*}

Given the language instruction $L$, coarse prior map $M_0^{prior}$, and onboard RGB/LiDAR observations, the system aligns geometric measurements with the prior in a unified world coordinate frame and maintains image-to-3D correspondences for subsequent grounding. On this shared representation, CoFiNav adopts a modular design that separates high-level semantic reasoning from geometric execution \cite{SayCan,SayPlan,VoxPoser}. The VLM progressively advances a hierarchical search and generates a structured observation request. The geometric module grounds this request into a 3D target, constructs the required observation constraints, and selects a feasible viewpoint before motion execution. Thus, the VLM determines \textbf{what to observe}, geometry determines \textbf{where to observe it}, and motion planning determines \textbf{how to reach it}, as illustrated in Fig.~\ref{fig:overview}.

\subsection{VLM-Guided Hierarchical Semantic Search}

We organize task-relevant entities, attributes, and relations into a dynamic hierarchical search structure, drawing on hierarchical 3D representations \cite{Dynamic3DSG,Hydra}. Let the search tree at semantic decision step $k$ be

\begin{equation}
\mathcal{T}_k=(\mathcal{N}_k,\mathcal{E}_k),
\end{equation}

where $\mathcal{N}_k$ denotes search nodes and $\mathcal{E}_k$ encodes task-dependent relationships. A node expands as new task-relevant observations become available.

Each node maintains a current search state \(s_i \in \{\text{unexplored}, \text{unresolved}, \text{confirmed}, \text{rejected}\}\). An unexplored node has not yet received an informative observation, whereas an unresolved node has been observed but remains ambiguous. Confirmed or rejected nodes have been verified or ruled out. Nodes that remain worth exploring form the frontier $\mathcal{F}_k\subseteq\mathcal{N}_k$, over which a priority queue allocates observations among competing facades, regions, and visual candidates.

At each semantic decision step, the VLM receives the language instruction, current RGB observation, image-aligned geometric context, and search history. The state of the current node is summarized by

\begin{equation}
D_k=(d_k^{vis},d_k^{id},d_k^{rel}),
\end{equation}

which respectively describe whether the target is sufficiently observable, identifiable, and adequate for verifying the required spatial relation. Insufficiency may therefore be associated with five observation factors: visibility, scale, angle, occlusion, and context. Together with task relevance, accumulated observations, remaining uncertainty, search history, and observation cost, this state is used to rank the frontier and select the next node:

\begin{equation}
n_k^{next}=\mathrm{VLM}(L,I_k,\mathcal{T}_k,\mathcal{F}_k,H_k,D_k).
\end{equation}

After selecting a node, the VLM generates the structured observation request

\begin{equation}
q_k=(n_k^{next},o_k,r_k^{\mathrm{obs}},c_k),
\end{equation}

where $o_k$ specifies the entity or spatial region to inspect, $r_k^{\mathrm{obs}}$ specifies the observation requirement (e.g., overview, larger projected scale, improved viewing direction, occlusion removal, or preserved context), and optional $c_k$ provides an image-space cue when the decision concerns a visible candidate. Targets already associated with explicit 3D support can be specified directly through $o_k$.

After a new observation is acquired, the tree and frontier are updated. Confirmed nodes may activate downstream dependencies, rejected nodes leave the active frontier, and unresolved nodes remain with revised priorities. Useful incidental observations are also stored and may update related nodes before they are explicitly visited. The search state evolves as

\begin{equation}
\mathcal{T}_{k+1},\mathcal{F}_{k+1}=
\mathrm{Update}(\mathcal{T}_k,\mathcal{F}_k,q_k,O_{k+1}).
\end{equation}

\subsection{Geometry-Constrained Observation Planning and Execution}

Given $q_k$, the geometric module resolves its semantic target into 3D space, translates the requested observation condition into geometric constraints, selects an executable viewpoint, and plans the corresponding motion.

\subsubsection{Geometric Support Construction}

Initial LiDAR observations acquired above the task region are aligned with the coarse prior \(M_0^{\mathrm{prior}}\) in the world coordinate frame to construct the geometric map \(M_{\mathrm{geo}}\).

The prior provides coarse structures such as building footprints and roads, while LiDAR complements them with observed surfaces, heights, and obstacles. For building regions, point clouds are associated with prior footprints to construct regularized geometric supports for facades and local subregions, including their world-coordinate boundaries, spatial extents, and surface orientations. We denote these supports by $\mathcal{S}_{geo}$. The underlying 3D map is retained for line-of-sight and safety checks, and subsequent LiDAR observations update local geometry during flight.

\subsubsection{Observation Target Resolution}

The requested target $o_k$ is resolved into a world-frame 3D support $\mathcal{R}_k$. Facades and structured surface regions can be retrieved or constructed from $\mathcal{S}_{geo}$, while previously localized candidates reuse their stored 3D support. When the request refers to a candidate in the current RGB image, the system grounds the corresponding image region through the image-to-world mapping

\begin{equation}
\phi_k:(u,v)\mapsto \mathbf{p}^w,
\end{equation}

where $(u,v)$ is an image coordinate and $\mathbf{p}^w$ is the corresponding 3D surface point. The target-resolution operation is expressed as

\begin{equation}
\mathcal{R}_k=\mathrm{Resolve}(o_k,\mathcal{S}_{geo},\phi_k,c_k).
\end{equation}

Depending on target type, $\mathcal{R}_k$ retains the spatial position, boundary, surface orientation, and other geometric information needed for viewpoint planning. This provides a common 3D interface for semantic targets originating from either the structured map or current visual observations.

\subsubsection{Constraint-Aware Viewpoint Planning}

The observation requirement $r_k^{\mathrm{obs}}$ is converted into geometric constraints $\Gamma_k$. A global overview requires sufficient field-of-view coverage; identification requires adequate projected scale; viewing-direction requests constrain the angle between the camera ray and target surface; occlusion requests require a new unobstructed line of sight; and relational verification requires the target and relevant reference region to remain jointly visible.

Based on the target position, extent, and surface orientation, the planner generates a task-conditioned candidate set \(\mathcal{V}_k=\{v_k^1,v_k^2,\ldots,v_k^m\}\) within an active-viewpoint planning formulation \cite{NBV}.

Each candidate contains UAV position and camera viewing direction. For explicit surfaces, candidates are sampled with respect to the target normal, tangent, and height direction; for local 3D targets, positions are generated according to target extent and surrounding free space.

The 3D map is then used to validate candidates. For sampled target point $\mathbf{p}_j$, ray casting determines its visibility from viewpoint $v$, and the weighted visibility ratio is

\begin{equation}
R_{\mathrm{vis}}(v)=
\frac{\sum_j w_j\,\mathrm{Vis}(\mathbf{p}_j,v)}
{\sum_j w_j},
\end{equation}

where $w_j$ weights task-relevant parts of the target. Candidates violating $\Gamma_k$, flight-safety, or reachability constraints are discarded, yielding $\mathcal{V}_k^{\mathrm{feas}}$. Among them, the final viewpoint balances observation quality and motion cost:

\begin{equation}
v_k^\ast=
\arg\max_{v\in\mathcal{V}_k^{\mathrm{feas}}}
\left[
Q_{\mathrm{obs}}(v,\Gamma_k,\mathcal{R}_k)
-\lambda C_{\mathrm{move}}(\xi_k,v)
\right].
\end{equation}

Here, $Q_{\mathrm{obs}}$ scores how well a candidate satisfies the current observation requirements, $C_{\mathrm{move}}$ denotes the motion cost from the current UAV pose, and $\lambda$ is a weighting coefficient that balances observation quality and motion cost.

\subsubsection{Motion Planning and Closed-Loop Execution}

Once \(v_k^\ast\) is selected, the motion planner computes a safe trajectory \(\tau_k\) from the current pose \(\xi_k\) to the selected observation viewpoint using \(M_{\mathrm{geo}}\). The path must satisfy collision and flight-safety constraints, while online LiDAR measurements support local feasibility checking and replanning when necessary. At the selected viewpoint, the UAV acquires a new RGB/LiDAR observation $O_{k+1}$ and returns it to the hierarchical search module. The complete closed loop is

\begin{equation}
q_k
\rightarrow
\mathcal{R}_k
\rightarrow
\mathcal{V}_k^{\mathrm{feas}}
\rightarrow
v_k^\ast
\rightarrow
\tau_k
\rightarrow
O_{k+1}.
\end{equation}

\section{EXPERIMENTS}

\subsection{Experimental Setup}

\subsubsection{Simulation Platform}

\begin{figure}[t]
    \centering
    \includegraphics[width=\columnwidth]{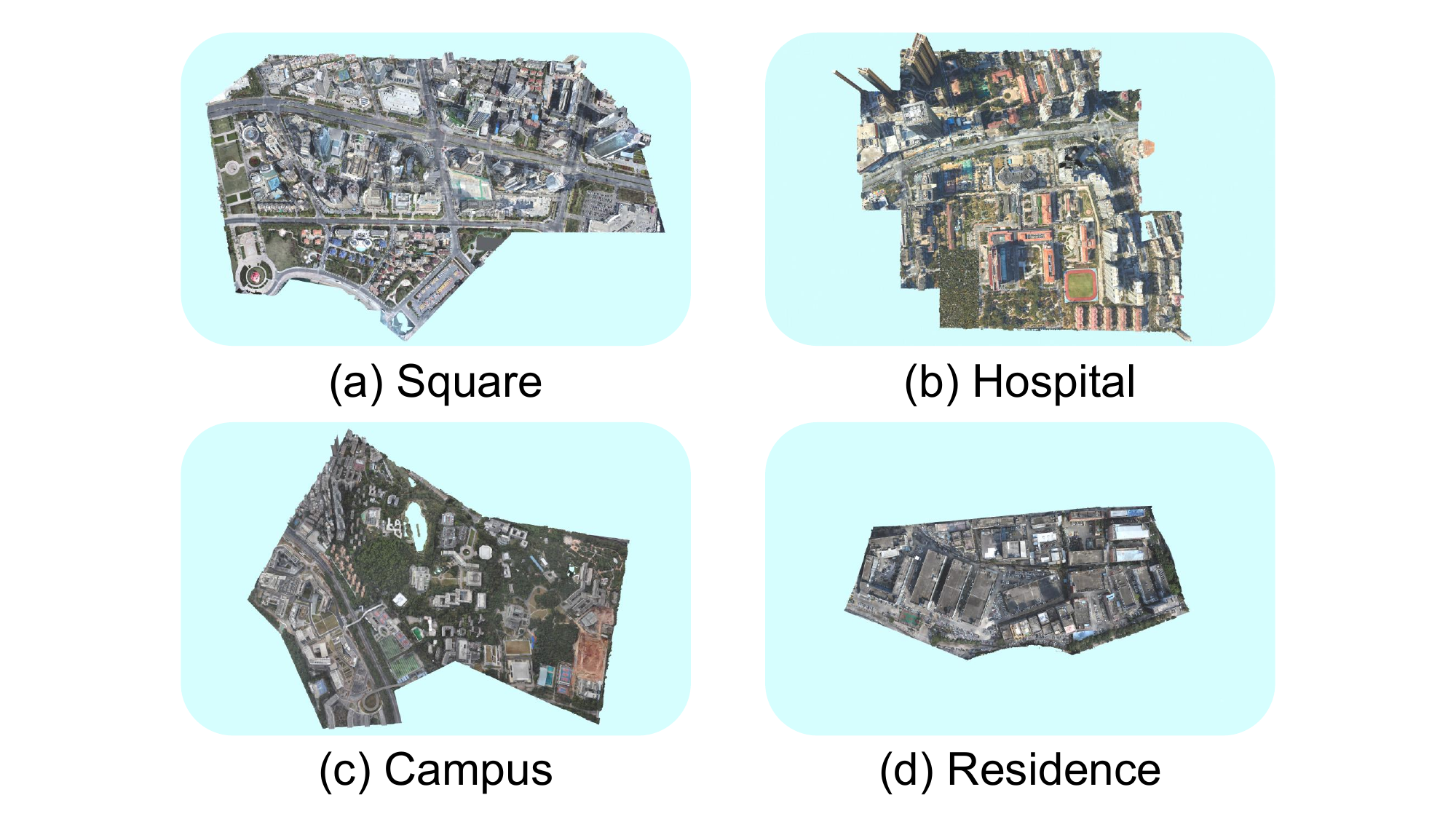}
    \caption{Overview of the simulation environments used for evaluation. Four UrbanScene3D scenes with different urban layouts are considered: (a) Square, (b) Hospital, (c) Campus, and (d) Residence.}
    \label{fig:simulation_environments}
\end{figure}

We construct the simulation platform based on UrbanScene3D \cite{UrbanScene3D}, using high-fidelity reconstructions of real urban scenes to preserve complex building geometry and rich visual details. We align the reconstructed layouts with corresponding OSM-derived coarse geographic priors, while keeping facade-level and local semantic information available only through onboard perception. As shown in Fig.~\ref{fig:simulation_environments}, we select four scenes---Square, Hospital, Campus, and Residence---to cover representative urban layouts and functions that support fine-grained terminal-navigation tasks.

\begin{figure*}[t]
    \centering
    \includegraphics[width=\textwidth]{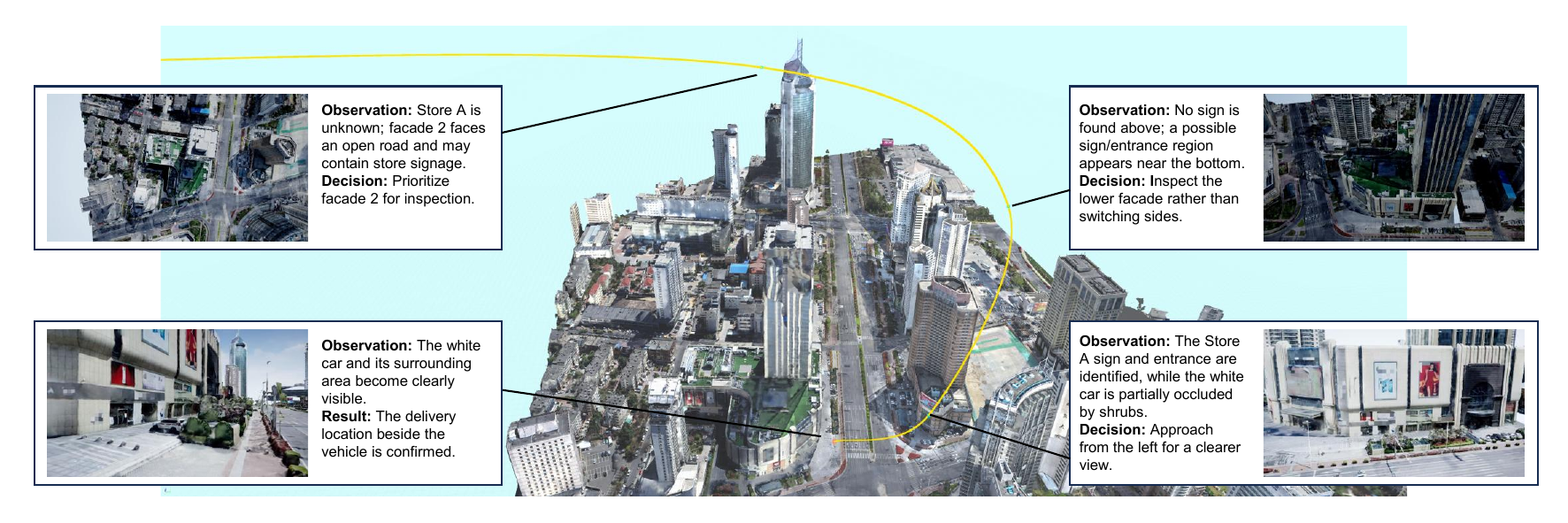}
    \caption{Representative coarse-to-fine navigation episode. The central aerial view shows the UAV approach trajectory, while the linked observations illustrate the progressive grounding process from facade selection and lower-facade inspection to entrance identification, occlusion-aware viewpoint adjustment, and final delivery-location confirmation beside the white car.}
    \label{fig:coarse_to_fine_episode}
\end{figure*}

We build the platform on Unreal Engine 4.27 with AirSim \cite{AirSim} and PX4. CoFiNav uses an RGB camera with adjustable viewing orientation and an omnidirectional 3D LiDAR for semantic observation, geometric grounding, viewpoint planning, and local safety checking.

On this platform, we construct 40 last-mile delivery tasks, with 10 tasks in each scene, and repeat every task for three independent trials, yielding 120 episodes for each evaluated configuration. All configurations use the same task set, initial poses, coarse geographic priors, flight regions, and evaluation criteria, while retaining the perception and search mechanisms required by each method.

\subsubsection{Evaluation Metrics}

An episode is successful only when the system terminates at a predicted target position within the corresponding ground-truth valid delivery region defined in Sec. III. Success Rate (SR) is computed over all 120 episodes. Path Length (PL) measures cumulative UAV travel distance, and Number of Semantic Observations (\#Obs) counts task-level visual observations that trigger an update of the navigation or search decision. Each episode is limited to 15 such observation steps, after which an unresolved task is terminated. PL and \#Obs characterize the physical and semantic search costs required under the same task-level evaluation budget.

\subsection{Comparison with Adapted Baselines}

To evaluate the performance of CoFiNav, we compare it with three representative systems discussed in Sec. II that are most closely related to our task, as described below. We retain the core navigation mechanisms of these approaches and make the adaptations required for evaluation on the same task.

\suppressfloats[t]
\begin{table}[t]
\caption{Comparison with adapted navigation and search baselines.}
\label{tab:baseline-comparison}
\centering
\footnotesize
\begin{tabular}{lccc}
\hline
Method & SR (\%) $\uparrow$ & PL (m) $\downarrow$ & \#Obs $\downarrow$ \\
\hline
GeoNav$^{\dagger}$ & 4.2 & 352.6 & 14.1 \\
VLFM$^{\dagger}$ & 13.3 & 406.8 & 12.2 \\
OPEN$^{\dagger}$ & 64.2 & 398.4 & 10.3 \\
CoFiNav & \textbf{81.7} & \textbf{286.4} & \textbf{6.6} \\
\hline
\end{tabular}
\end{table}

\subsubsection{GeoNav$^{\dagger}$}

GeoNav$^{\dagger}$ adapts GeoNav \cite{GeoNav} to our task while retaining its original fixed-altitude, top-down aerial observation regime. The UAV can relocate horizontally to obtain different aerial views of the target region, but does not descend to facade-level or local close-range viewpoints. This adaptation retains GeoNav's hierarchical geospatial reasoning while applying its original observation paradigm to fine-grained terminal grounding.

\subsubsection{VLFM$^{\dagger}$}

We use the UAV-adapted VLFM implementation \cite{VLFM} released with ESARBench \cite{ESARBench} while retaining the original VLFM search pipeline and target-scoring mechanism. The complete language instruction is provided directly to the search agent. The coarse building prior is used only to define a broad local search region around the task building, preventing exploration from drifting to unrelated parts of the city without prescribing specific viewpoints or search order.

\subsubsection{OPEN$^{\dagger}$}

OPEN was originally developed for ground-robot last-mile address navigation \cite{OpenBench}. We adapt its local exploration procedure to UAV flight by generating viewpoints around the target building and visiting them according to its uniformly sampled coverage strategy. After the relevant terminal cue is identified, we additionally provide the final grounding stage required by our task so that the adapted baseline can produce a relation-constrained delivery position. This strengthens the original pipeline for full-task evaluation while preserving its core coverage-based observation strategy.

\begin{figure}[t]
    \centering
    \includegraphics[width=\columnwidth]{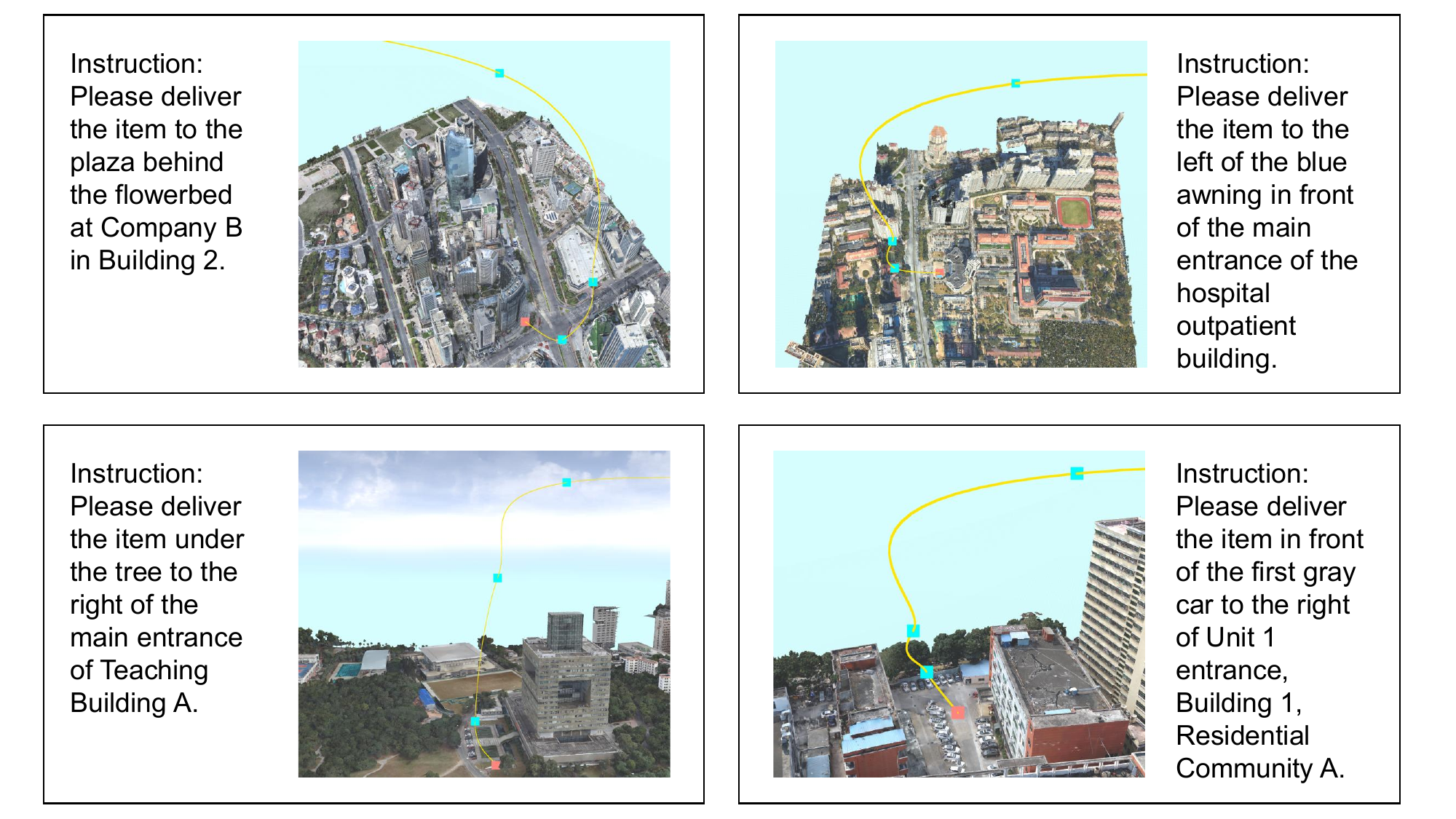}
    \caption{Representative CoFiNav trajectories across the four evaluation environments. Each panel shows one language-specified delivery task and its corresponding UAV trajectory. Cyan markers indicate semantic observation viewpoints, the yellow curve denotes the executed trajectory, and the red marker indicates the final grounded delivery location.}
    \label{fig:representative_trajectories}
\end{figure}

Quantitative results are reported in Table~\ref{tab:baseline-comparison}. A representative CoFiNav episode is illustrated in Fig.~\ref{fig:coarse_to_fine_episode}, while Fig.~\ref{fig:representative_trajectories} shows trajectories across the four evaluation environments.

CoFiNav achieves the highest SR of 81.7\%, together with the shortest PL of 286.4 m and the fewest semantic observations at 6.6. GeoNav$^{\dagger}$ and VLFM$^{\dagger}$ achieve only 4.2\% and 13.3\% SR, respectively, while both require substantially longer flight paths and more semantic observations. Even after being enhanced by final relational grounding, OPEN$^{\dagger}$ reaches 64.2\% SR and still requires 398.4 m and 10.3 semantic observations on average.

CoFiNav adapts 3D viewpoints to the observation requirement, providing the scale and viewing geometry needed for fine-grained terminal grounding. GeoNav$^{\dagger}$ retains a fixed-altitude and top-down observation regime, where facade-level cues often remain too small and horizontal relocation cannot consistently recover the required viewing geometry, resulting in a low SR. CoFiNav maintains task-relevant entities and spatial relations throughout hierarchical semantic search, while VLFM$^{\dagger}$ often selects semantically plausible candidates without preserving the spatial context required by the instruction, leading to grounding errors and local convergence on incorrect candidates. OPEN$^{\dagger}$ uses uniformly sampled building-perimeter exploration. It evaluates many unrelated viewpoints and cannot actively adjust projected scale. This sometimes produces partial observations of task-relevant entities and spatial relations, increasing path length and observation cost while limiting accuracy. CoFiNav instead uses the evolving semantic state to determine which information should be acquired next and grounds the request into a task-conditioned viewpoint, achieving a higher SR with shorter flight paths and fewer observations.

\begin{figure}[t]
    \centering
    \includegraphics[width=\columnwidth]{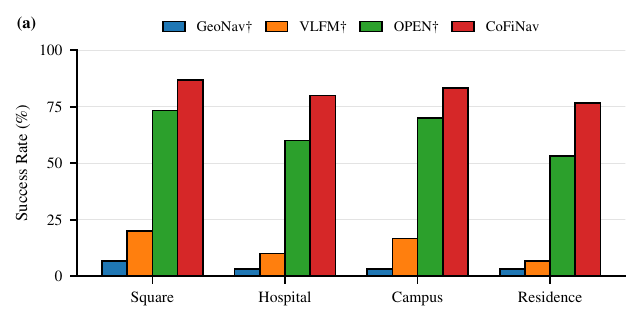}
    \includegraphics[width=\columnwidth]{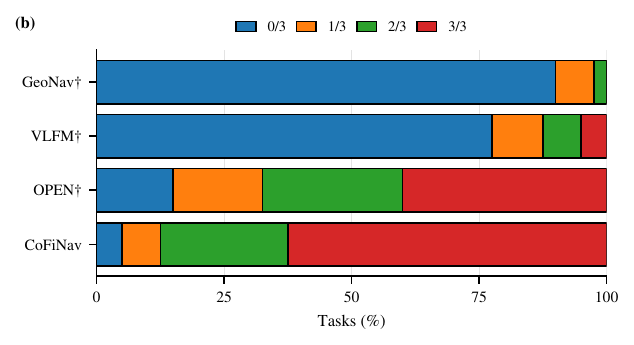}
    \caption{Performance consistency across evaluation environments and tasks. (a) Scene-wise success rate across the four UrbanScene3D environments. (b) Distribution of tasks according to the number of successful outcomes across three independent trials; 0/3--3/3 denote the number of successful trials for each task.}
    \label{fig:performance_consistency}
\end{figure}

To further examine whether the aggregate advantage in Table~\ref{tab:baseline-comparison} is concentrated in particular environments or tasks, Fig.~\ref{fig:performance_consistency} provides a more detailed breakdown of the comparison. Fig.~\ref{fig:performance_consistency}(a) reports scene-wise success rates across Square, Hospital, Campus, and Residence, while Fig.~\ref{fig:performance_consistency}(b) groups the 40 tasks according to the number of successful trials among three independent trials. The scene-wise results show that CoFiNav maintains its performance advantage across different urban environments. The task-level distribution further shows that its aggregate success rate is supported by a larger proportion of tasks that are completed consistently across repeated trials.

\begin{table}[t]
\caption{Ablation results of CoFiNav.}
\label{tab:ablation-v39}
\centering
\begin{tabular}{lccc}
\hline
Variant & SR (\%) $\uparrow$ & PL (m) $\downarrow$ & \#Obs $\downarrow$ \\
\hline
w/o Hierarchical State & 43.3 & 411.2 & 11.4 \\
w/o Semantic Priority & 64.2 & 376.5 & 10.2 \\
w/o Viewpoint Optimization & 49.2 & 321.7 & 8.6 \\
CoFiNav & \textbf{81.7} & \textbf{286.4} & \textbf{6.6} \\
\hline
\end{tabular}
\end{table}

\begin{table}[t]
\caption{Comparison across vision-language models.}
\label{tab:vlm-v39}
\centering
\begin{tabular}{lccc}
\hline
VLM & SR (\%) $\uparrow$ & PL (m) $\downarrow$ & \#Obs $\downarrow$ \\
\hline
GPT-5 & \textbf{81.7} & 286.4 & \textbf{6.6} \\
Qwen3-VL-235B-A22B-Thinking & 79.2 & 294.8 & 6.9 \\
Gemini 2.5 Pro & 75.8 & \textbf{281.7} & 7.1 \\
Qwen3-VL-Flash & 71.7 & 308.6 & 7.6 \\
\hline
\end{tabular}
\end{table}

\subsection{Ablation Study}

We investigate three key components of CoFiNav. \textbf{w/o Hierarchical State} removes the persistent coarse-to-fine organization of resolved and unresolved search states, leaving candidate observations to be explored without hierarchical progression. \textbf{w/o Semantic Priority} retains the hierarchy and node states but selects active frontier nodes according to their insertion order rather than VLM-based ranking. \textbf{w/o Viewpoint Optimization} retains hierarchical search, target resolution, candidate generation, and safety checks, but removes task-conditioned optimization of projected scale, viewing angle, occlusion, and contextual visibility.

As shown in Table~\ref{tab:ablation-v39}, removing the hierarchical state produces the largest disruption to the search process. Without persistent coarse-to-fine organization, observations can shift between locally plausible candidates without maintaining consistent progress toward the terminal goal, leading to both lower SR and substantial search cost. Removing semantic priority retains the hierarchical structure and therefore preserves a relatively high SR, but substantially increases \#Obs and PL because unresolved candidates are no longer visited according to their task relevance. Removing viewpoint optimization also causes a pronounced SR reduction, as the selected poses may provide line of sight without sufficient scale, viewing direction, or contextual visibility for reliable grounding. These results show complementary roles for hierarchical progression, semantic prioritization, and task-conditioned viewpoint selection.

\subsection{Comparison Across Vision-Language Models}

We evaluate CoFiNav with GPT-5, Qwen3-VL-235B-A22B-Thinking, Gemini 2.5 Pro, and Qwen3-VL-Flash. As shown in Table~\ref{tab:vlm-v39}, CoFiNav remains effective across all four VLMs, with SR ranging from 71.7\% to 81.7\%. GPT-5 achieves the highest SR and the fewest semantic observations, while Gemini 2.5 Pro yields the shortest PL. The variation across model backends indicates that semantic reasoning quality affects search robustness, while the shared hierarchical, geometric, and motion components preserve the overall coarse-to-fine navigation behavior.

\subsection{Failure Analysis}

Among the remaining failures, the most common cases occur in visually ambiguous terminal scenes containing repetitive facades or multiple similar candidates, where VLM-level semantic errors can affect candidate identification or relational interpretation. A smaller number of failures are associated with severe occlusion or difficult geometric grounding during close-range observation. Low-level flight failures are comparatively uncommon, indicating that the navigation and execution pipeline remains generally stable once the relevant terminal cues are correctly resolved. These cases suggest that further performance gains may primarily come from improved fine-grained semantic discrimination and relational grounding under challenging visual conditions.

\section{CONCLUSION}

This paper presented CoFiNav, a VLM-guided coarse-to-fine framework for UAV terminal semantic navigation and last-mile delivery. CoFiNav treats terminal localization as a progressive grounding process in which unresolved semantic entities and spatial relations guide the acquisition of task-relevant observations during approach. A hierarchical semantic search mechanism determines what information should be resolved next, while geometry-grounded active 3D observation converts these semantic requirements into feasible viewpoints for safe execution. By progressively resolving terminal destinations from onboard observations during approach, CoFiNav extends UAV delivery beyond predefined terminal specifications toward more flexible open-ended semantic descriptions. Experiments across multiple urban environments show that CoFiNav outperforms representative adapted baselines in success rate while requiring shorter flight paths and fewer observations. Future work will focus on further improving the robustness and reliability of VLM-based grounding in complex urban environments.

\IEEEtriggeratref{20}

\end{document}